\documentclass{article}

\PassOptionsToPackage{numbers,sort&compress}{natbib}
 \usepackage[preprint]{neurips_2026}
\usepackage[utf8]{inputenc} 
\usepackage[T1]{fontenc}    
\usepackage{hyperref}       
\usepackage{url}            
\usepackage{booktabs}       
\usepackage{amsfonts}       
\usepackage{nicefrac}       
\usepackage{microtype}      
\usepackage{xcolor}         
\usepackage{xcolor}
\usepackage[utf8]{inputenc} 
\usepackage{booktabs}       
\usepackage{amsfonts}       
\usepackage{microtype}      
\usepackage{algpseudocode}
\usepackage{algorithm}
\usepackage[nolist]{acronym}
\usepackage{verbatim}
\usepackage{colortbl}
\usepackage{arydshln}
\usepackage{graphicx}
\usepackage{amsmath}
\usepackage{wrapfig}

\usepackage{booktabs,tabularx}
\usepackage{siunitx} 
\usepackage{adjustbox} 
\usepackage{subcaption}  
\usepackage{makecell}

\begin{acronym}[DRAM]
    \acro{AI}{Artificial Intelligence}
    \acrodefindefinite{AI}{an}{an}
    \acro{NN}{Neural Network}
    \acro{DL}{Deep Learning}
    \acro{ML}{Machine Learning}
    \acro{TM}{Tsetlin Machine}
    \acro{NLP}{Natural Language Processing}
    \acro{LLMs}{Large Language Models}
    \acro{LLM}{Large Language Model}
    \acro{TM-AE}{Tsetlin Machine Autoencoder}
    \acro{RbE}{Reasoning by Elimination}
    \acro{TW}{Target Word}
    \acro{CoTM}{Coalesced Tsetlin Machine}
    \acro{Omni TM-AE}{Omni Tsetlin Machine Autoencoder}
    \acro{FastOmniTMAE}{Fast Omni Tsetlin Machine Autoencoder}
    \acrodefindefinite{ML}{an}{a}
    \acro{TA}{Tsetlin automaton}
    \acrodefplural{TA}{Tsetlin automata}
    \acro{TAT}{Tsetlin automaton team}
    \acro{GPT}{Generative Pre-trained Transformer}
    \acro{EDA}{Easy Data Augmentation}
    \acro{PLM}{Pre-Trained Language Model}
    \acro{PLMs}{Pre-Trained Language Models}
    \acro{GPU}{Graphics Processing Unit}
    \acro{CPU}{Central Processing Unit}
    \acro{FPGA}{Field-Programmable Gate Array}
    \acro{IoT}{Internet of Things}
    \acro{CUDA}{Compute Unified Device Architecture}
    \acro{SM}{Streaming Multiprocessor}
    \acrodefplural{SM}{Streaming Multiprocessors}
    \acro{SoC}{System on Chip}
    \acro{AXI}{Advanced eXtensible Interface}
    \acro{PS}{Processing System}
    \acro{PL}{Programmable Logic}
    \acro{IP}{Intellectual Property}
\end{acronym}

\title{FastOmniTMAE: Parallel Clause Learning for Scalable and Hardware-Efficient Tsetlin Embeddings}
\author{%
  Ahmed K.~Kadhim\\
  Department of ICT\\
  University of Agder\\
  Grimstad, Norway\\
  \texttt{ahmed.k.kadhim@uia.no} \\
  \And
  Lei Jiao\\
  Department of ICT\\
  University of Agder\\
  Grimstad, Norway\\
  \texttt{lei.jiao@uia.no} \\
  \And
  Rishad Shafik\\
  School of Engineering\\
  Newcastle University\\
  Newcastle upon Tyne, UK\\
  \texttt{rishad.shafik@newcastle.ac.uk} \\
  \And
  Ole-Christoffer Granmo\\
  Department of ICT\\
  University of Agder\\
  Grimstad, Norway\\
  \texttt{ole.granmo@uia.no} \\
  \And
  Mayur Kishor Shende\\
  Department of ICT\\
  University of Agder\\
  Grimstad, Norway\\
  \texttt{mayur.kishor.shende@uia.no} \\
}

\begin{document}
    \maketitle
    \begin{abstract}
Embedding models in natural language processing (NLP) increasingly rely on deep architectures such as BERT, while simpler models such as Word2Vec provide efficient representations but limited interpretability. The Tsetlin Machine (TM) offers an alternative logic-based learning paradigm. Omni TM Autoencoder (Omni TM-AE) applies this paradigm to static embedding by exploiting automaton state distributions within a single clause layer, but its training process remains slow.
In this work, we propose FastOmniTMAE, a reformulation of Omni TM-AE that replaces sequential training dependencies with a two-stage parallel process: evaluation and update. Using a Single-Run Multi-Environment Benchmark covering classification, similarity, and clustering, FastOmniTMAE achieves up to 5$\times$ faster training in classification while maintaining comparable embedding quality under both Spearman and Kendall similarity measures.
To address the limited efficiency of TM training on conventional GPUs, we further implement FastOmniTMAE as a reusable accelerator on SoC-FPGA platforms. The Multi-Hardware Benchmark shows that FastOmniTMAE achieves similarity scores of 0.669 on a resource-constrained FPGA and 0.696 on an UltraScale+ SoC, demonstrating efficient logic-based embedding training with a small hardware footprint.
\end{abstract}


    \section{Introduction}
\ac{AI} models are rapidly evolving and becoming increasingly complex to meet the demand for higher accuracy and broader capabilities. Among the most prominent examples are large language models such as ChatGPT and Gemini, which are primarily based on transformer architectures. However, these models have grown to a scale where understanding and verifying their decision-making processes has become challenging. This limitation arises from their reliance on neural networks, where knowledge is represented through continuous numerical values, making reasoning difficult to interpret. Consequently, there is a growing need for alternative approaches that provide transparency and explainability, particularly in applications where understanding the reasoning process is essential.

The \ac{TM} offers one such alternative for building transparent \ac{AI} models~\citep{tm}. It is inherently interpretable in both structure and reasoning. Structurally, \ac{TM} is composed of logical operations governed by \acp{TA}, which follow a well-defined learning mechanism~\citep{ta-hierarchical}. From a reasoning perspective, decisions are formed through clauses, which represent logical expressions capturing relevant patterns in the data~\cite{clause-size}. These clauses are directly traceable and interpretable, as demonstrated in prior works \citep{jeeru2025interpretable, tm-interpretable-rules}. 
Another key advantage of \ac{TM} is its hardware-friendly nature, stemming from its reliance on simple logical operations. Previous studies \cite{parallel-tm} and \cite{cmos-tm} have explored hardware implementations of \ac{TM}, primarily focusing on inference for classification tasks, with limited work addressing training on hardware~\citep{tm-edge,asic}. This design enables the development of \ac{AI} systems with a small computational footprint, without requiring specialized architectures typically used for neural network acceleration~\citep{on-chip-tm,edge-inference}.

Previous work on \ac{TM-AE}~\citep{tm-ae,scalable-tmae,rbe-ahmed,adversarial}, including the \ac{Omni TM-AE} model~\citep{omni}, has primarily focused on achieving high embedding quality in comparison with models such as Word2Vec, GloVe, BERT, and FastText~\citep{word2vec,glove,devlin2018bert,fastText}. However, training efficiency has received limited attention, despite being a significant limitation. Reported training times can extend to several months~\cite{scalable-tmae}, while \ac{Omni TM-AE} reduces this to several days by eliminating multi-layer structures~\cite{omni}. 
In this work, we analyze the \ac{TM} architecture with a focus on training efficiency across different hardware platforms, highlighting its limitations, particularly on \acp{GPU}. We then introduce a redesigned version of the \ac{Omni TM-AE} model, termed \ac{FastOmniTMAE}, which improves training speed while preserving embedding quality. To evaluate embedding performance, we develop a dedicated benchmark that systematically assesses embedding quality in \ac{NLP} tasks. This model also serves as the foundation for exploring training performance across heterogeneous hardware platforms.

The main contributions of this work are summarized as follows:
\begin{enumerate}
    \item Developing a new \ac{Omni TM-AE} architecture, \ac{FastOmniTMAE}, that improves training speed while maintaining comparable embedding quality.
    \item Introducing a benchmark (Single-Run Multi-Environment Benchmark) for evaluating embedding quality in \ac{NLP} applications.
    \item Analyzing the effectiveness of \ac{GPU}-based training for \ac{TM} and identifying associated inefficiencies.
    \item Demonstrating the first full-scale training of a \ac{TM}-based embedding model on \ac{FPGA} hardware for \ac{NLP}.
    \item Proposing a benchmark (Multi-Hardware Benchmark) for comprehensive comparison across \acp{CPU}, \acp{GPU}, and \ac{FPGA}-based systems.
\end{enumerate}

    \section{Methodology}

In this section, we first describe the \ac{Omni TM-AE} approach in detail to highlight the weaknesses of the algorithm. We then present the proposed \ac{FastOmniTMAE} method. Finally, we discuss the main challenges of implementing \ac{TM} structures on hardware and how the proposed design addresses these challenges.

\subsection{The Tsetlin Machine and Embedding Architecture}

\begin{figure}
    \centering
    \includegraphics[width=0.9\linewidth]{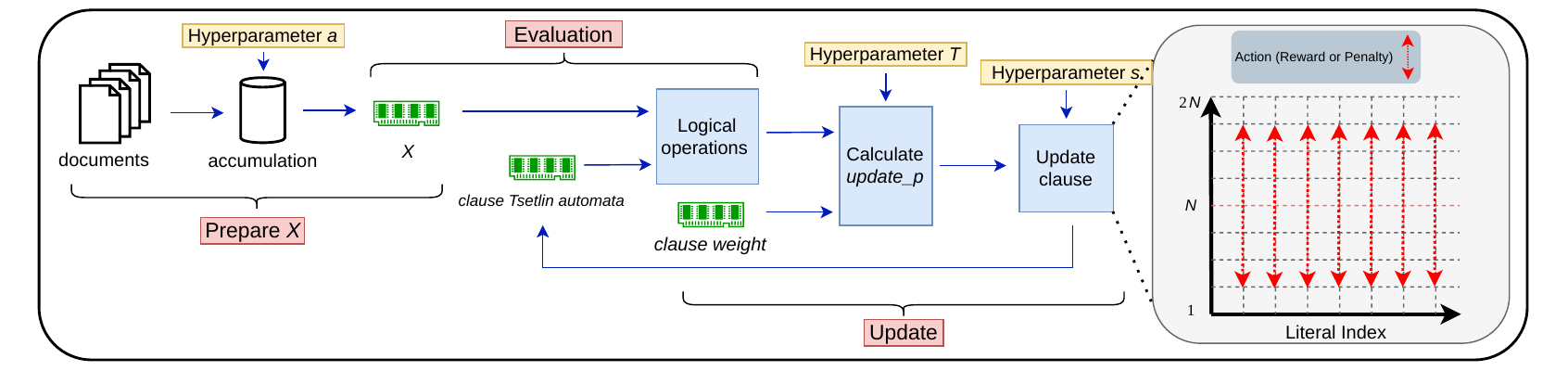}
    \caption{Overview of the \ac{TM}-based embedding training flow, showing input accumulation, clause evaluation, update-probability calculation, and automaton-state updates.}
    \label{fig:tm_training_overview}
\end{figure}

The operation of the \ac{TM} is primarily based on basic learning units \acp{TA}.
In \ac{NLP} applications, each automaton corresponds to a feature in the vocabulary. For example, given a vocabulary $(x_1, x_2, x_3)$, each feature is assigned an automaton that captures its learned state through training.
Automata are combined into logical expressions called clauses, which represent class-specific patterns. 
Multiple clauses may represent the same class, and clauses may also be shared across classes, as in the \ac{CoTM} structure~\citep{cotm}. Figure~\ref{fig:tm_training_overview} summarizes this flow, from document accumulation and clause evaluation to update-probability calculation and automaton-state updates.

When feature negation is used, the clause space is extended from features to literals, e.g., $(x_1, x_2, x_3, \neg x_1, \neg x_2, \neg x_3)$. 
Each automaton operates in a discrete state space controlled by \(b\) state bits, giving \(2^b\) possible states. 
In compact binary storage, each automaton requires \(b\) bits. Thus, for \(C\) clauses and $L$ literals, the automata state memory is \(M_{\mathrm{TA}} = C \times L \times b\) bits. 
For example, when \(b=8\), each automaton has 256 possible states, ranging from 0 to 255. An intermediate threshold \(N\) determines whether a literal is included in clause evaluation. Literals with states above \(N\) are included, while those below \(N\) are excluded. 
For a single clause \(C_j\), the clause output $c_j$ that is used for evaluation is therefore the logical $AND$ of its included literals \(l\):
{\small
\[
c_j = \bigwedge_{l \in L,\ \mathrm{state}(l) > N} l .
\]
}
For example, a clause representing the target word “space” may include literals such as “vehicle”, “vacuum”, and “light” during evaluation, while still retaining states for all literals.

The first stage of clause training is evaluation, where logical operations such as \(XOR\) and \(AND\) compare the input example \(X\) with the clause structure. A supportive clause evaluation reinforces selected automata, while a non-supportive evaluation penalizes them. State updates are performed through increment or decrement operations. During training, each clause undergoes three main update types: (i) memorizing operation, which reinforces true literals in \(X\) when the clause agrees with the input and the output is 1, while false literals may undergo probabilistic forgetting controlled by the hyperparameter \(s\); (ii) forgetting operation, which decreases the influence of true literals when the clause does not agree with the input, with probability governed by \(s\); and (iii) invalidation operation, which reinforces false literals to modify the clause and reject incorrect patterns.

In addition to the automata memory state matrix, the \ac{CoTM} structure includes a weights matrix. This matrix assigns integer weights to clauses, allowing them to be promoted or suppressed during voting. 

The \ac{Omni TM-AE} model is built on the \ac{CoTM} structure combined with an autoencoder framework~\citep{omni}. Unlike traditional autoencoders, \ac{Omni TM-AE} is trained separately for each target token, enabling the reuse of generated embedding vectors. It also uses the full automaton state space, including excluded literals below the threshold \(N\). This allows the model to extract embeddings directly from the clause states without requiring an additional embedding extraction stage.
As a result, \ac{Omni TM-AE} produces static embedding vectors similar to Word2Vec and GloVe. 

Despite its effectiveness, \ac{Omni TM-AE} suffers from long training times, reaching several minutes per token on large datasets, which limits its practicality for large-scale applications. For each target token, the \ac{Omni TM-AE} training procedure consists of the following stages:
\begin{enumerate}
    \item Output preparation: Define a binary label ($output$), where 1 indicates the presence of the target token, and 0 indicates its absence from documents in the dataset.
    \item Input preparation: Construct \(X\) by randomly accumulating \(a\) documents according to the target token label.
    \item Evaluation: Evaluate each clause output $c_j$ against the input \(X\) to produce a binary clause output $o_j$, indicating whether the included literals of clauses are satisfied.
    \item Class sum calculation: Aggregate the weighted clause outputs, clip the result by the hyperparameter \(T\), and compute \(update\_p\), which defines the probability of updating clauses:
    {\small
        \[
        class\_sum = \mathrm{clip}\left(\sum_{j=1}^{C} w_j o_j,\,-T,\,T\right),
        \qquad
        update\_p =
        \begin{cases}
        \dfrac{T - class\_sum}{2T}, & output = 1, \\[3pt]
        \dfrac{T + class\_sum}{2T}, & output = 0.
        \end{cases}
        \]
    }
    Here, \(j\) indexes each clause in \(C\), and \(w_{j}\) is the weight of clause $C_j$.
    \item Clause update: Update clauses according to the output and weights.
\end{enumerate}


\subsubsection{Class Sum Calculation Analysis}
\label{apdx:class_sum}

\begin{figure}
    \centering
    \includegraphics[width=1\linewidth]{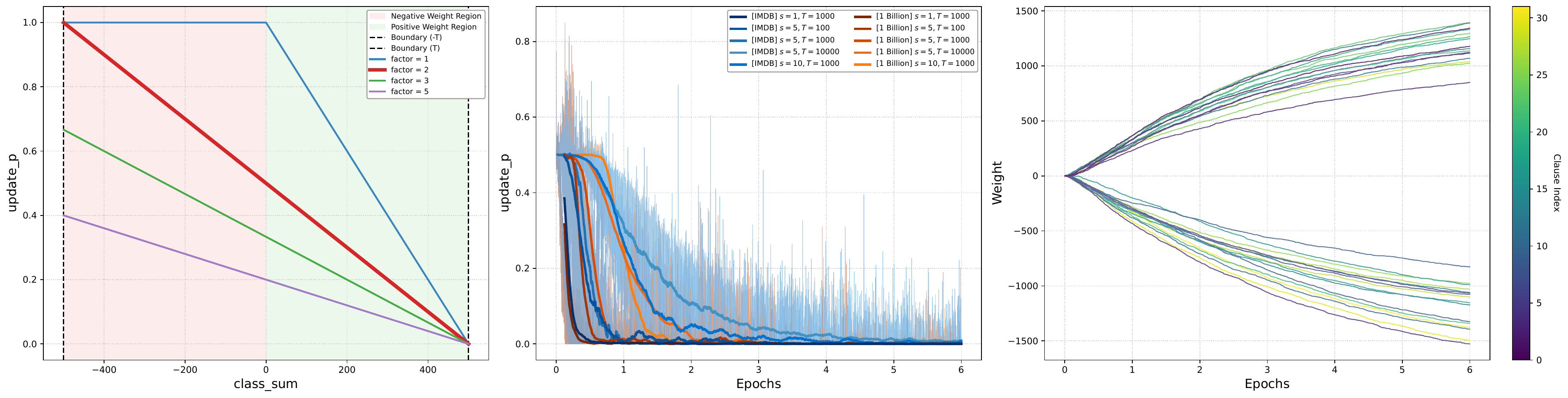}
    \caption{(a on the left) Relationship between the $class\_sum$ and the update probability $update\_p$. (b in the middle) Update probability $update\_p$ over epochs for different datasets and hyperparameters. (c on the right) Clause weights variation over epochs.}
    \label{fig:class_sum_updae_p}
\end{figure}

One of the key components of the \ac{TM} training process is the $class\_sum$ calculation, which is typically used to determine the voting outcome for the outputs. In the case of \ac{Omni TM-AE}, where only two outputs are considered  (0 and 1), the primary role of $class\_sum$ is to compute the update rate $update\_p$ and guide the convergence of the model.

Specifically, $class\_sum$, and ultimately $update\_p$, controls the proportion of clauses selected for updating. This mechanism plays an important role in driving the model toward a stable state, where clauses are sufficiently trained to represent the underlying patterns and no longer require significant updates. At this stage, training can be terminated.

Figure~\ref{fig:class_sum_updae_p}(a) illustrates the relationship between $class\_sum$ and $update\_p$. The vertical axis represents the $update\_p$, while the horizontal axis corresponds to $class\_sum$. The figure includes multiple curves reflecting different scaling factors in the update probability equation (e.g., division by 2 in the formulation above). In practice, a factor of 2 is commonly used, as it provides a balanced adjustment between negatively weighted clauses (red region) and positively weighted clauses (blue region).

Since \(class\_sum\) is bounded by \(T\), the factor-2 curve shows a neutral update rate of approximately 50\% when neither positive nor negative clauses dominate. Activation of many negative clauses increases the update rate above 50\%, while activation of many positive clauses reduces it below 50\%, indicating that the clauses are becoming stable and require fewer updates.
\subsection{FastOmniTMAE}
In this work, the \ac{Omni TM-AE} model is redesigned to improve training speed while maintaining comparable embedding quality. Two main limitations in the original training procedure were identified.

The first limitation is the $class\_sum$ calculation. This step requires all clauses to be evaluated before any update can be performed, creating a global dependency in the training process. As discussed previously, this mechanism is intended to guide convergence~\citep{convergence-tm}. While this may be suitable for small-scale applications, it becomes inefficient in \ac{NLP} tasks with large vocabularies and many training documents.

Figure~\ref{fig:class_sum_updae_p}(b) illustrates this behavior on two datasets: IMDb~\cite{imdb}, with fewer than 2,000 tokens, and the 1 Billion Word dataset~\cite{one_billion}, with 40,000 tokens. The curves are generated using different values of \(s \in \{1,5,10\}\) and \(T \in \{100,1000,10000\}\), where each epoch contains 2,000 training examples. At the beginning of training, the update rate is approximately 50\%, since no clauses have yet adapted to the input patterns. As training progresses, $class\_sum$ increases and the update rate gradually decreases toward zero.

This convergence behavior does not necessarily imply sufficient training for high-quality embeddings. For example, IMDb provides limited contextual information, leading to weak embeddings regardless of convergence. In contrast, large-scale datasets provide richer contextual diversity, allowing stronger embeddings to be obtained within a small number of epochs. Thus, embedding quality in \ac{TM}-based NLP models is more strongly influenced by data richness than by convergence behavior alone.

The second limitation concerns the weight matrix. In general \ac{CoTM} formulations~\citep{cotm}, weights support voting across multiple classes. However, in \ac{Omni TM-AE}, where only binary outputs are used, weights mainly influence $class\_sum$ and therefore act as a convergence-scaling mechanism. As shown in Figure~\ref{fig:class_sum_updae_p}(c), weights follow a monotonic pattern during training, with positive and negative weights preserving their signs. This indicates that their main role is to reduce the update rate rather than alter the learned decision structure.

Based on these observations, explicit $class\_sum$ calculation is not required for embedding quality. Instead, training for a sufficient number of epochs is adequate. Removing $class\_sum$ eliminates global synchronization across clauses and reduces evaluation waiting time.

To address this bottleneck, we redesign the training process by shifting from global to local decision-making. Clause evaluation and updating are decoupled and performed independently, with the update probability computed locally from each clause’s evaluation outcome. This enables parallel execution of evaluation and update operations and forms the foundation of the \ac{FastOmniTMAE} model. The classification and similarity results in the Single-Run Multi-Environment Benchmark show that removing the global \(class\_sum\) dependency does not degrade embedding quality compared with \ac{Omni TM-AE}, but primarily removes a training bottleneck. A detailed description of the \ac{FastOmniTMAE} training procedure is provided in Appendix~\ref{app:fastomni_algorithm}.

\subsubsection{Convergence Interpretation}

Through extensive experimentation and the developed benchmarks used to rebuild the model while maintaining high embedding quality, it became necessary to identify reliable indicators of convergence and determine when further training is no longer required. A potential approach is to analyze the distribution of literals within the automaton state space.

As discussed in~\citet{omni,rbe-ahmed}, as the vocabulary size and training data increase, obtaining a clear distribution becomes more difficult. With a small number of epochs, the model tends to push irrelevant literals toward lower states, while important literals involved in clause formation are elevated toward the upper threshold ($N$). This process typically relies on feature negation to create a distribution within the negated literal space, which is then used in embedding computation, as in the \ac{Omni TM-AE} model.

Although this distribution may appear to indicate sufficient training, it does not necessarily correlate with embedding quality. At present, there is no established relationship between the state distribution and embedding performance. Identifying such a relationship could enable early stopping criteria and the formulation of a convergence measure tailored to \ac{TM}-based embedding models. 
The observed state distributions are provided in Appendix~\ref{app:state_distribution}.

\subsubsection{TM on Hardware}

The proposed \ac{FastOmniTMAE} approach demonstrates efficient performance on \acp{CPU} due to its parallel structure. An implementation was also developed using \ac{CUDA} for \ac{GPU} execution. However, the observed performance was limited compared to deep learning models, even on high-end systems such as the NVIDIA DGX H100.

Modern \acp{GPU}, such as those based on the Hopper H100 architecture, are designed to accelerate floating-point computations. Each \ac{GPU}  consists of multiple \acp{SM}, each containing \ac{CUDA} cores, Tensor cores, and additional specialized units. These components are optimized for operations such as fused multiply-add (FMA), which are fundamental to neural network computations~\citep{nvidia_hopper_architecture}.

In contrast, the \ac{TM} primarily relies on simple logical operations, including $AND$, $OR$, $NOT$, $Shift$ operations, and $XOR$. While these operations are computationally lightweight, they are not efficiently supported by \ac{GPU}  architectures optimized for dense floating-point arithmetic. A significant portion of \ac{GPU}  hardware resources is dedicated to FMA-based computation, while only a fraction is allocated to integer-based operations. Consequently, much of the \ac{GPU}  architecture remains underutilized when executing \ac{TM}-based workloads.

Additionally, specialized units such as Tensor Cores and Ray Tracing Cores are not applicable to \ac{TM} operations, leading to further inefficiencies. These observations highlight the mismatch between \ac{TM} computational requirements and \ac{GPU}  design, motivating the need for dedicated hardware solutions to efficiently accelerate \ac{TM} training.

\subsubsection{The New Hardware Design}

The proposed design shifts the computational burden of \ac{FastOmniTMAE} training from general-purpose processors to a dedicated accelerator implemented on \ac{FPGA} hardware. This approach addresses the limitations of sequential execution and leverages the logical nature of the \ac{TM} architecture.

The accelerator is implemented as a reusable \ac{IP} core, with variants targeting both resource-constrained \ac{SoC} platforms, such as Zynq-7000, and higher-performance \ac{SoC}-FPGA platforms, such as Zynq UltraScale+ MPSoC. The design adopts a batch-processing strategy that enables autonomous and continuous training while minimizing dependence on the host processor.

\begin{figure}
    \centering
    \includegraphics[width=0.9\linewidth]{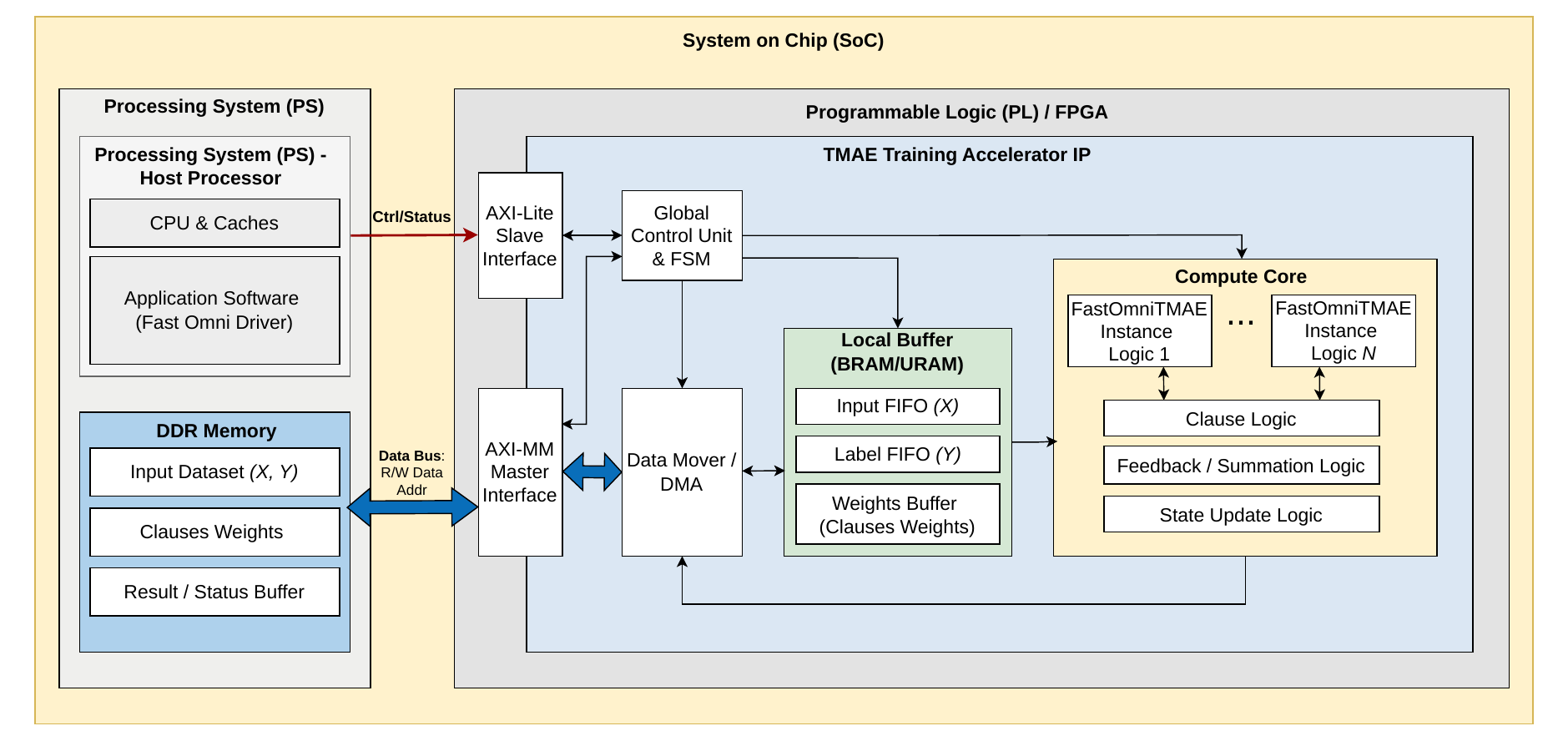}
    \caption{SoC-based architecture of the \ac{FastOmniTMAE} accelerator with PS–PL communication over AXI interfaces for efficient parallel training.}
    \label{fig:fpga}
\end{figure}

Figure\ref{fig:fpga} illustrates the system architecture. The \ac{SoC} platform consists of a \ac{PS} and a \ac{PL} component (\ac{FPGA}). Communication between these components is achieved using standard \ac{AXI} protocols. Two primary interfaces are used:
\begin{itemize}
    \item AXI-Lite: A control interface that allows the processor to configure training parameters, including memory addresses, batch size, and control signals such as start and reset.
    \item AXI-MM (Memory-Mapped): A high-throughput interface that enables direct memory access for reading training data and updating automata states stored in external DDR memory, minimizing processor involvement and reducing latency.
\end{itemize}

The design adopts a two-stage parallel training mechanism (evaluation and update), ensuring both efficiency and correctness. Training data is streamed from external memory into on-chip buffers (BRAM) to reduce communication overhead.

The accelerator is controlled by a finite state machine (FSM) that manages the training workflow:
\begin{enumerate}
    \item Data Fetching: Loading input features $X$ and labels $Y$ in batches.
    \item Logical Computation: Executing \ac{TM} operations directly using logic gates, resulting in lower power consumption compared to floating-point operations on \acp{GPU}.
    \item State Update: Updating automata states and writing results back to external memory for subsequent iterations.
\end{enumerate}

The design is scalable and adaptable to different hardware constraints. A Zybo board implementation targets resource-constrained \ac{SoC} platforms (Zynq-7000), prioritizing minimal logic utilization and efficient memory access, while a ZCU104 board implementation exploits the higher bandwidth and larger programmable-logic resources of the UltraScale+ MPSoC to support greater parallelism and larger-scale training. Additional implementation details are provided in the appendix.
    \section{Evaluation}

\subsection{Single-Run Multi-Environment Benchmark}

The purpose of the first benchmark was to validate the development stages of the proposed \ac{FastOmniTMAE} model through continuous comparison with the original \ac{Omni TM-AE} model. The experiments were conducted on an NVIDIA DGX H100 server with 8 \acp{GPU} and 2.0 TiB of RAM. DevContainer was used to create two isolated environments for \ac{FastOmniTMAE} and \ac{Omni TM-AE} to ensure a fair comparison. The benchmark included three tests: classification, similarity, and clustering. The additional baseline models used in the evaluation were executed in separate environments. The benchmark code will be released publicly after double-blind review.

\subsubsection{Classification Performance}
\label{sec:classification}

\begin{figure}
    \centering
    \includegraphics[width=0.9\linewidth]{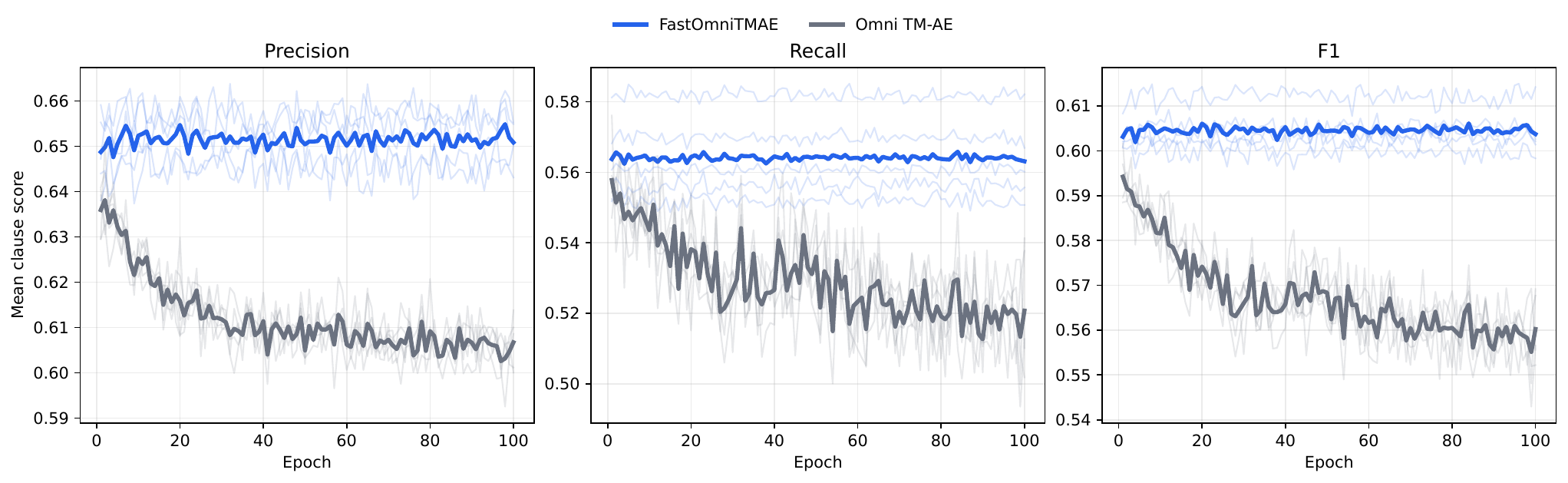}
    \caption{Classification performance of \ac{FastOmniTMAE} and \ac{Omni TM-AE} over 100 epochs using precision, recall, and F1-score.}
    \label{fig:classification}
\end{figure}

For the classification task, 100 tokens were selected. Training was performed on a 1 Billion Words dataset using 1,000 examples per token, followed by 500 examples for inference. The dataset vocabulary contained 40k tokens, and each token’s documents were labeled as True or False. True indicates that the document supports the target token, while False indicates that the document does not contain the target token. Figure~\ref{fig:classification} shows the precision, recall, and F1-score for \ac{FastOmniTMAE} and \ac{Omni TM-AE}. Each experiment was repeated 5 times, with 100 epochs per run. The 100 target tokens are a configurable benchmark subset and can be expanded to larger or full-vocabulary runs.

In Figure~\ref{fig:classification}, the dark curve represents the mean over paired runs, while the lighter curves in the background represent individual runs. 
Both models were trained using identical settings: $clauses=160$, $T=8000$, $s=2$, and $seed=42$. 
\ac{FastOmniTMAE} demonstrated stable performance throughout training, while \ac{Omni TM-AE} showed a gradual decline as training progressed. \ac{FastOmniTMAE} consistently achieved F1-scores above 0.6 across epochs, whereas \ac{Omni TM-AE} started below 0.6 and decreased to 0.56 in the final epochs. This decline is attributed to the use of $class\_sum$ to reduce the clause update rate, which limits the model’s ability to adapt to previously unseen patterns. As discussed earlier, this behavior does not provide a reliable convergence measure and may prevent continued learning when further updates are needed. In addition to its higher classification performance, \ac{FastOmniTMAE} achieved more than a \textbf{5×} speedup. \ac{Omni TM-AE} required 1709 seconds to train for 100 epochs, while \ac{FastOmniTMAE} completed the same training in 310 seconds. This demonstrates the effectiveness of the proposed model structure in improving both training speed and accuracy.
\subsection{Semantic Similarity Evaluation}
\label{sec:similarity}

\begin{table*}[t]
\caption{Word-similarity benchmark results for embeddings. Entries report the mean and standard deviation over five runs; \(\rho\) denotes Spearman correlation and \(\tau\) denotes Kendall correlation.}
\label{tab:similarity}
\centering
\resizebox{\textwidth}{!}{%
\begin{tabular}{@{}llcccccc@{}}
\toprule
\textbf{Model} & \textbf{Metric} & \textbf{RG-65} & \textbf{WS-353-Sim} & \textbf{MTurk-287} & \textbf{MTurk-771} & \textbf{MEN} & \textbf{Avg.} \\
\midrule
GloVe & $\rho$ & 0.528 {\scriptsize $\pm$ 0.013} & 0.514 {\scriptsize $\pm$ 0.005} & 0.521 {\scriptsize $\pm$ 0.004} & 0.461 {\scriptsize $\pm$ 0.001} & 0.564 {\scriptsize $\pm$ 0.001} & 0.518 {\scriptsize $\pm$ 0.004} \\
 & $\tau$ & 0.365 {\scriptsize $\pm$ 0.010} & 0.355 {\scriptsize $\pm$ 0.004} & 0.365 {\scriptsize $\pm$ 0.004} & 0.319 {\scriptsize $\pm$ 0.000} & 0.394 {\scriptsize $\pm$ 0.001} & 0.360 {\scriptsize $\pm$ 0.003} \\
\addlinespace[2pt]
Word2Vec & $\rho$ & 0.529 {\scriptsize $\pm$ 0.006} & 0.591 {\scriptsize $\pm$ 0.002} & 0.558 {\scriptsize $\pm$ 0.004} & 0.471 {\scriptsize $\pm$ 0.002} & 0.561 {\scriptsize $\pm$ 0.002} & 0.542 {\scriptsize $\pm$ 0.001} \\
 & $\tau$ & 0.357 {\scriptsize $\pm$ 0.007} & 0.415 {\scriptsize $\pm$ 0.002} & 0.389 {\scriptsize $\pm$ 0.003} & 0.324 {\scriptsize $\pm$ 0.002} & 0.391 {\scriptsize $\pm$ 0.002} & 0.375 {\scriptsize $\pm$ 0.001} \\
\addlinespace[2pt]
FastText & $\rho$ & 0.508 {\scriptsize $\pm$ 0.007} & 0.601 {\scriptsize $\pm$ 0.003} & 0.574 {\scriptsize $\pm$ 0.004} & 0.486 {\scriptsize $\pm$ 0.003} & 0.580 {\scriptsize $\pm$ 0.002} & 0.550 {\scriptsize $\pm$ 0.001} \\
 & $\tau$ & 0.348 {\scriptsize $\pm$ 0.007} & 0.420 {\scriptsize $\pm$ 0.002} & 0.399 {\scriptsize $\pm$ 0.004} & 0.334 {\scriptsize $\pm$ 0.002} & 0.406 {\scriptsize $\pm$ 0.001} & 0.382 {\scriptsize $\pm$ 0.002} \\
\addlinespace[2pt]
\ac{Omni TM-AE} & $\rho$ & 0.633 {\scriptsize $\pm$ 0.020} & 0.478 {\scriptsize $\pm$ 0.013} & 0.519 {\scriptsize $\pm$ 0.006} & 0.487 {\scriptsize $\pm$ 0.004} & 0.596 {\scriptsize $\pm$ 0.002} & 0.543 {\scriptsize $\pm$ 0.005} \\
 & $\tau$ & 0.452 {\scriptsize $\pm$ 0.016} & 0.344 {\scriptsize $\pm$ 0.012} & 0.367 {\scriptsize $\pm$ 0.005} & 0.350 {\scriptsize $\pm$ 0.003} & 0.439 {\scriptsize $\pm$ 0.002} & 0.390 {\scriptsize $\pm$ 0.004} \\
\addlinespace[2pt]
FastOmniTMAE (CPU) & $\rho$ & 0.656 {\scriptsize $\pm$ 0.016} & 0.470 {\scriptsize $\pm$ 0.007} & 0.498 {\scriptsize $\pm$ 0.009} & 0.477 {\scriptsize $\pm$ 0.003} & 0.586 {\scriptsize $\pm$ 0.002} & 0.537 {\scriptsize $\pm$ 0.003} \\
 & $\tau$ & 0.468 {\scriptsize $\pm$ 0.014} & 0.343 {\scriptsize $\pm$ 0.006} & 0.356 {\scriptsize $\pm$ 0.009} & 0.350 {\scriptsize $\pm$ 0.001} & 0.441 {\scriptsize $\pm$ 0.001} & 0.392 {\scriptsize $\pm$ 0.003} \\
\addlinespace[2pt]
FastOmniTMAE (CUDA) & $\rho$ & 0.649 {\scriptsize $\pm$ 0.020} & 0.469 {\scriptsize $\pm$ 0.014} & 0.492 {\scriptsize $\pm$ 0.006} & 0.472 {\scriptsize $\pm$ 0.006} & 0.576 {\scriptsize $\pm$ 0.002} & 0.532 {\scriptsize $\pm$ 0.008} \\
 & $\tau$ & 0.465 {\scriptsize $\pm$ 0.015} & 0.347 {\scriptsize $\pm$ 0.009} & 0.352 {\scriptsize $\pm$ 0.005} & 0.347 {\scriptsize $\pm$ 0.004} & 0.434 {\scriptsize $\pm$ 0.002} & 0.389 {\scriptsize $\pm$ 0.006} \\
\bottomrule
\end{tabular}%
}
\end{table*}

In the second test, we used the similarity datasets RG65, MTurk287, MTurk771, WordSim353, and MEN. These datasets contain human-annotated word pairs, where each pair is assigned a similarity score. Table~\ref{tab:similarity} compares \ac{FastOmniTMAE} with other models included in the benchmark, namely Word2Vec, FastText, GloVe, and \ac{Omni TM-AE}. For \ac{Omni TM-AE} and \ac{FastOmniTMAE}, the same hyperparameters were used: $number\_of\_examples=2000$, $accumulation = 24$, $clauses=32$, $T=20000$, $s=1.0$, $epochs=4$, and $number\_of\_state\_bits=8$. The remaining models were trained using $vector\_size=100$, $window=5$, and $epochs=25$. All models were trained on the \ac{CPU} using the same DGX H100 server, except that \ac{FastOmniTMAE} was also trained on \ac{GPU} to evaluate the suitability of \ac{GPU} acceleration for TM-based training.

From Table~\ref{tab:similarity}, FastText achieved the highest average Spearman correlation ($\rho=0.550$), followed closely by \ac{FastOmniTMAE} on CPU ($\rho=0.537$). In terms of Kendall correlation, \ac{FastOmniTMAE} achieved the highest average score ($\tau=0.392$), slightly outperforming \ac{Omni TM-AE} ($\tau=0.390$) and FastText ($\tau=0.382$). \ac{FastOmniTMAE} also achieved the highest individual score on the RG65 dataset for both Spearman and Kendall correlations, representing the best single-dataset similarity result in the experiment. Overall, \ac{FastOmniTMAE} achieved comparable embedding quality while improving training efficiency over \ac{Omni TM-AE} on CPU under identical resource allocation. Both models were trained using approximately 80\% of the DGX H100 CPU resources, corresponding to around 179 cores. \ac{FastOmniTMAE} required an average of 242.9 seconds, while \ac{Omni TM-AE} required 293.1 seconds, giving \ac{FastOmniTMAE} a \textbf{1.2$\times$} speedup on the same hardware. The training times of the remaining models are not directly compared, as they were executed under different resource conditions. For GPU execution, \ac{FastOmniTMAE} showed longer training time due to the mismatch between GPU architecture and TM-based logical operations. This issue is discussed further in the Multi-Hardware Benchmark section.

\subsection{Clustering Analysis}
\label{sec:clustering}

\begin{wrapfigure}{r}{0.5\textwidth}
    \centering
    \vspace{-10pt}
    \includegraphics[width=0.48\textwidth]{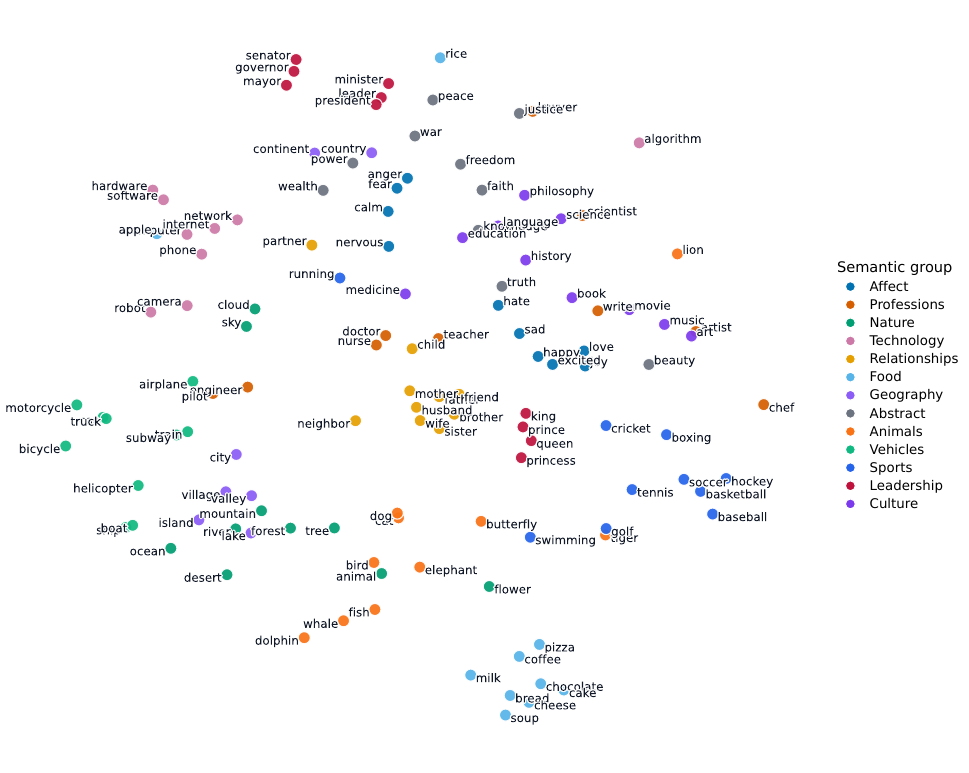}
    \caption{t-SNE visualization of \ac{FastOmniTMAE} embeddings for pre-defined semantic word groups.}
    \label{fig:tsne}
    \vspace{-10pt}
\end{wrapfigure}

The previous tests provide structured numerical evaluation. The final test in this benchmark provides a visual representation of embedding quality. In this test, t-SNE was used to cluster a set of words based on their embedding vectors. The selected words were pre-divided into distinguishable semantic groups, such as food, geography, and vehicles.
The embeddings used for this visualization were generated with 32 clauses, \(T=20000\), \(s=1.0\), accumulation \(=24\), 4 epochs, and 2000 training examples per token.
Figure~\ref{fig:tsne} shows the resulting clusters, demonstrating the ability of \ac{FastOmniTMAE} to generate meaningful embedding vectors. Words belonging to the same semantic group are generally clustered together, indicating that the model captures useful semantic relationships.

\subsection{ Multi-Hardware Benchmark}

\begin{table*}[t]
\caption{\ac{FastOmniTMAE} hardware benchmark on the RG65 44-token workload (352k examples total). Each row reports the mean with standard deviation over 5 runs.}
\label{tab:hardware}
\centering
\resizebox{\linewidth}{!}{%
\begin{tabular}{@{}llcccc@{}}
\toprule
Platform & Configuration & Backend & Time (s) $\downarrow$ & Speedup vs. local \ac{CPU} $\uparrow$ & RG65 $\rho$ $\uparrow$ \\
\midrule
ZCU104 Board & ARM CPU (1 of 4 cores) & CPU & 848.7 {\scriptsize$\pm$ 1.5} & 1.00$\times$ & 0.680 \\
ZCU104 Board & FPGA design (12 engines) & SoC & 119.9 {\scriptsize$\pm$ 0.9} & 7.08$\times$ & 0.696 {\scriptsize$\pm$ 0.005} \\
\midrule
Zybo Board & ARM CPU (1 of 2 cores) & CPU & 3667.5 {\scriptsize$\pm$ 21.8} & 1.00$\times$ & 0.680 \\
Zybo Board & FPGA design (2 engines) & SoC & 657.4 {\scriptsize$\pm$ 1.1} & 5.58$\times$ & 0.669 {\scriptsize$\pm$ 0.008} \\
\midrule
Core i5 PC & Host CPU (1 of 8 threads) & CPU & 137.3 {\scriptsize$\pm$ 37.1} & 1.00$\times$ & 0.680 \\
Core i5 PC & Intel Iris Xe Graphics & OpenCL & 475.6 {\scriptsize$\pm$ 7.2} & 0.29$\times$ & 0.668 \\
\midrule
Core i9 PC & Host CPU (1 of 24 threads) & CPU & 57.7 {\scriptsize$\pm$ 1.9} & 1.00$\times$ & 0.680 \\
Core i9 PC & Intel Graphics & OpenCL & 262.2 {\scriptsize$\pm$ 0.4} & 0.22$\times$ & 0.668 \\
\midrule
H100 Server & Host CPU (1 of 224 threads) & CPU & 87.3 {\scriptsize$\pm$ 0.8} & 1.00$\times$ & 0.680 \\
H100 Server & NVIDIA H100 80GB HBM3 & CUDA & 493.1 {\scriptsize$\pm$ 0.5} & 0.18$\times$ & 0.681 \\
\bottomrule
\end{tabular}
}
\end{table*}

This benchmark compares the performance of the same \ac{FastOmniTMAE} model across different hardware architectures. \ac{FastOmniTMAE} was implemented using multiple technologies to target \acp{CPU}, \acp{GPU}, and SoC-FPGA platforms. Accordingly, four configurations were developed: a C implementation for \ac{CPU} execution, an HDL implementation for FPGA acceleration, a C/OpenCL implementation for Intel \ac{GPU} execution, and a \ac{CUDA} implementation for NVIDIA \ac{GPU} execution. These configurations were tested on five devices: a mid-range Intel Core i5 PC, an Intel Core i9 Ultra workstation, an NVIDIA DGX H100 server, a Xilinx ZCU104 board, and a Digilent Zybo board.

To ensure a fair comparison, the benchmark was limited to training only, excluding example preparation. Preparing examples requires a corpus larger than 4 GB, which is unsuitable for constrained hardware such as the Zybo board with 1 GB of RAM. Therefore, examples for each token were prepared as small files containing 8,000 examples per token. Training consisted of two stages: clause evaluation and clause update. The examples were extracted from a 1 Billion Word dataset with a vocabulary of 40,000 entries. The RG65 word similarity dataset, which contains 44 word pairs, was selected as the evaluation task. This dataset was chosen based on previous benchmark observations, as it provides sufficient sensitivity to evaluate embedding quality. The hyperparameters were set as follows: 32 clauses, \(T=20000\), and \(s=1.0\).

Table~\ref{tab:hardware} presents the training results across the different hardware configurations. The fastest training time was achieved using a single-threaded implementation on the Intel Core i9 Ultra processor operating at 3.7 GHz. 
However, on the same machine, execution on the integrated \ac{GPU} reduced performance by a factor of 0.22. The DGX H100 server, with \ac{CPU} frequencies ranging from 2.0 to 2.9 GHz during training\footnote{Although the processor can reach up to 3.9 GHz under specific conditions, this mode was not used in our experiments.}, also achieved strong \ac{CPU} performance. 
In contrast, CUDA execution on the DGX H100 did not provide the expected acceleration, despite full \ac{GPU} utilization. As discussed earlier, this is attributed to the mismatch between \ac{GPU} architectures and the logical operation structure of the \ac{TM}. The Core i5 PC achieved moderate \ac{CPU} performance and acceptable \ac{GPU} performance, with better \ac{GPU} training time than the DGX H100 CUDA implementation.

Apart from \ac{CPU} execution, the best training performance was achieved on the ZCU104 SoC-FPGA board. This platform provided sufficient resources to instantiate 12 \ac{FastOmniTMAE} cores, allowing clauses to be trained in parallel. In addition to efficient execution time, this configuration achieved a high similarity score of \textbf{0.696}, making it the best-performing hardware configuration in terms of embedding quality. The design was also executable on the more resource-constrained Zybo board, demonstrating the scalability of the proposed architecture. Despite its limited resources, the Zybo implementation achieved a competitive similarity score with a smaller hardware footprint. Further details on hardware resource utilization for the ZCU104 and Zybo implementations are provided in Appendix~\ref{app:resource_utilization}.

    \section{Conclusion}

This work presented \ac{FastOmniTMAE}, a parallel reformulation of \ac{Omni TM-AE} that removes global training dependencies and improves training efficiency while preserving comparable embedding quality. Across classification, similarity, and clustering evaluations, \ac{FastOmniTMAE} demonstrated stable performance and faster training than \ac{Omni TM-AE}. The SoC-FPGA implementation further showed that TM-based embedding can benefit from dedicated logic-based hardware, achieving efficient training with a small hardware footprint.
    
    \small
    \bibliographystyle{plainnat}
    \bibliography{references}
    
    \appendix
    \section{Technical appendices and supplementary material}
    \subsection{Related Work}

Most hardware-oriented \ac{TM} research has focused on classification tasks, particularly in low-power and edge-computing settings. Early implementations demonstrated the suitability of \acp{TM} for constrained platforms such as microcontrollers and batteryless sensing systems~\citep{stm-arm-tm,TIMSP430-mcu-tm,batteryless-mcu-tm}. Other studies explored \ac{TM} execution on \acp{FPGA}, asynchronous hardware, and mixed software--hardware systems, showing that the logical structure of \acp{TM} can be mapped efficiently to diverse hardware targets~\citep{cyclone-fpga-iot-tm,fpga-synth-tm,risc-v-fpga-tm,software-fpga-tm,mcu_uart_tm}.

More recent work has investigated specialized \ac{TM} accelerators for image classification. In particular, convolutional and coalesced variants have been implemented on \acp{FPGA} and custom digital hardware, including designs with on-device training and highly energy-efficient inference~\citep{soc-fpga-tm,asic,cmos-tm}. These works demonstrate the hardware efficiency of \acp{TM}, but they primarily target image classification and do not address embedding generation or \ac{NLP} representation learning.

In \ac{NLP}, previous \ac{TM} work has mostly focused on classification and interpretability, while embedding-oriented models remain comparatively limited. \ac{Omni TM-AE} introduced a static embedding model based on the \ac{CoTM} structure, where clauses are trained per token and the full automaton state space is used to construct embeddings~\citep{cotm,omni}. Unlike conventional static embedding models such as Word2Vec and GloVe~\citep{word2vec,glove}, \ac{Omni TM-AE} derives representations directly from clause-state distributions. However, prior \ac{TM-AE} work has mainly emphasized embedding quality rather than training efficiency or hardware execution.

This work differs from previous hardware and embedding studies by targeting full-scale \ac{TM}-based embedding training for \ac{NLP}. Instead of accelerating inference for classification, we redesign the \ac{Omni TM-AE} training procedure and evaluate \ac{FastOmniTMAE} across CPUs, GPUs, and \ac{SoC}-FPGA platforms.
    \subsection{State-Space Distribution of Literals}
\label{app:state_distribution}

\begin{figure}[th]
    \centering
    \includegraphics[width=1\linewidth]{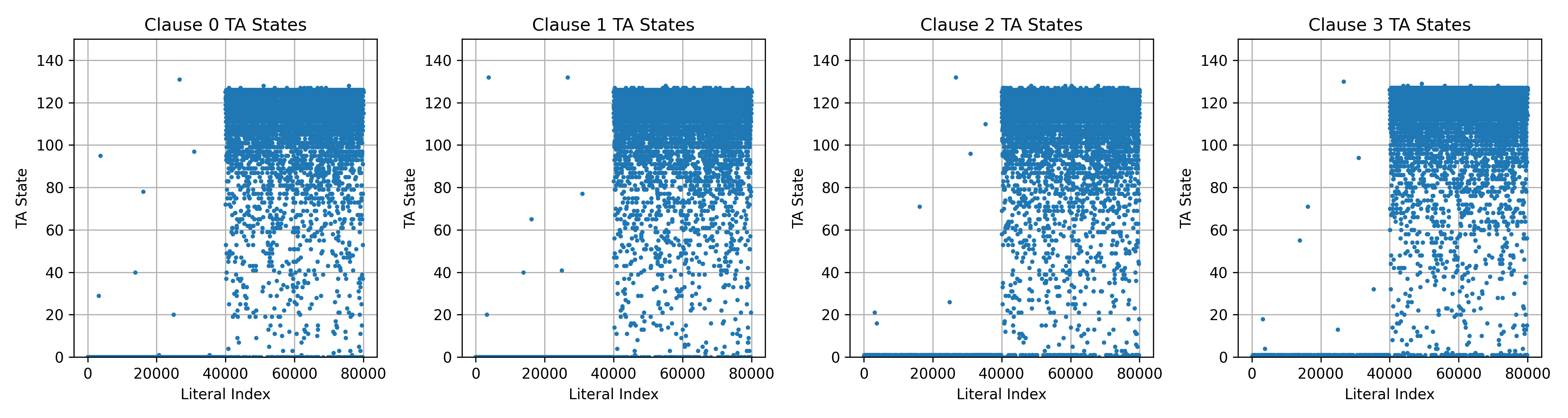}
    \caption{Distribution of \ac{TA} states across clauses after training. The first half of the literal index range corresponds to original vocabulary features, while the second half corresponds to negated literals. The distribution shows that original literals are mostly pushed toward low states, whereas negated literals form a richer high-state distribution, indicating that absence-based information is important for Omni-style embedding construction.}
    \label{fig:ta_state_distribution}
\end{figure}

Figure~\ref{fig:ta_state_distribution} visualizes the distribution of automaton states after training \ac{FastOmniTMAE}. Each subplot corresponds to one clause, where the horizontal axis represents the literal index and the vertical axis represents the learned \ac{TA} state. Since the vocabulary contains 40{,}000 original features, the full literal space contains 80{,}000 literals after including negations. Thus, the first half of the horizontal axis corresponds to original literals, while the second half corresponds to their negated counterparts.

The visualization shows that the model rapidly separates original and negated literals in the state space. Most original literals are pushed toward low states, indicating that they are not selected for clause evaluation. In contrast, the negated literals form a dense distribution across higher states, with many automata approaching or exceeding the inclusion threshold. This behavior suggests that the model does not rely only on the direct presence of words, but also learns from their absence. In other words, negated literals provide a form of reasoning by elimination, allowing clauses to characterize a target token not only by what supports it, but also by what should be excluded.

This observation is important for embedding construction. In conventional \ac{TM} usage, only literals whose automata exceed the inclusion threshold contribute directly to the clause expression. However, in \ac{Omni TM-AE} and \ac{FastOmniTMAE}, the full automaton state space is informative. Even excluded literals retain trained state values, and these values reflect how each feature interacts with the target token during training. Therefore, both included and excluded literals contribute to the final embedding representation through their signed state values.

The strong distribution in the negated literal region also explains why feature negation is important in the Omni embedding formulation. If only original literals were used, much of the learned information would be discarded, especially in large-vocabulary settings where many original features are suppressed. By combining the state values of original literals with the corresponding negated literals, the embedding vector captures a more complete representation of the target token.

However, the distribution should not be interpreted as a direct measure of embedding quality. Although a clear separation between original and negated literals indicates that the model has learned structured clause states, there is currently no established mapping between a specific state-distribution pattern and downstream embedding performance. This supports the observation made in the main text: state-space structure can provide useful insight into training dynamics, but it cannot yet serve as a reliable convergence criterion or early-stopping rule.

\subsection{FastOmniTMAE Training Procedure}
\label{app:fastomni_algorithm}

Algorithm~\ref{alg:fast_omni_tmae} summarizes the training procedure used in \ac{FastOmniTMAE}. 
The key difference from the original \ac{Omni TM-AE} training procedure is that clause evaluation and clause update are performed locally at the clause level. Instead of computing a global \textit{class\_sum} over all clauses before updating, each clause uses its own weighted contribution to determine its update probability. This removes the global synchronization step and allows clauses to be evaluated and updated independently.

For each target token, a binary autoencoder example is generated by sampling whether the target token is present or absent. The input vector \(X\) is constructed by accumulating documents according to this target label. Each clause \(C_j\) is then evaluated against \(X\), producing a binary output \(o_j\). The local clause contribution \(r_j = w_j o_j\) is used to compute the update probability. If selected for update, the clause receives one of the standard feedback types: Type~Ia reinforcement, Type~Ib penalization, or Type~II false-positive suppression. This procedure preserves the original \ac{TM} feedback mechanism while enabling parallel clause-level training~\cite{tm}.

\begin{algorithm}[t]
\caption{\ac{FastOmniTMAE} training procedure}
\label{alg:fast_omni_tmae}
\begin{algorithmic}[1]
\Require Training data \(\mathcal{D}\), target tokens \(\mathcal{Y}\), clauses \(\mathcal{C}\), weights \(W\), threshold \(T\), specificity \(s\), accumulation \(a\)
\For{each training iteration}
    \For{each target token \(y \in \mathcal{Y}\)}
        \State Sample binary target label \(Y \in \{0,1\}\)
        \State Construct input \(X\) by accumulating \(a\) documents according to \(Y\)
        
        \For{each clause \(C_j \in \mathcal{C}\)}
            \State Evaluate clause output:
            \[
            o_j \leftarrow \textsc{EvaluateClause}(X, C_j)
            \]
            \State Compute local clause contribution:
            \[
            r_j \leftarrow \mathrm{clip}(w_j o_j, -T, T)
            \]
            \State Compute update probability:
            \[
            p_j \leftarrow
            \begin{cases}
            \dfrac{T-r_j}{2T}, & Y=1, \\[4pt]
            \dfrac{T+r_j}{2T}, & Y=0 .
            \end{cases}
            \]
            
            \If{\(\textsc{SampleUpdate}(p_j)\) and \(w_j \geq 0\)}
                \If{\(Y=1\)}
                    \If{\(o_j=1\)}
                        \State Apply Type~Ia feedback:
                        \State \hspace{0.4cm} reinforce literals active in \(X\)
                        \State \hspace{0.4cm} penalize inactive literals
                    \Else
                        \State Apply Type~Ib feedback:
                        \State \hspace{0.4cm} penalize clause states
                    \EndIf
                \Else
                    \If{\(o_j=1\)}
                        \State Apply Type~II feedback:
                        \State \hspace{0.4cm} reinforce literals absent from \(X\) to suppress false positives
                    \EndIf
                \EndIf
            \EndIf
        \EndFor
    \EndFor
\EndFor
\end{algorithmic}
\end{algorithm}

\subsection{Resource Utilization for ZCU104 and Zybo}
\label{app:resource_utilization}

Table~\ref{tab:resource_utilization} summarizes the post-synthesis resource utilization of the proposed \ac{FastOmniTMAE} accelerator on the two \ac{SoC}-FPGA targets. The ZCU104 design targets the Zynq UltraScale+ MPSoC device \texttt{xczu7ev-ffvc1156-2-e} and instantiates 12 accelerator engines. The design uses a 150 MHz PL clock and distributes memory traffic across multiple high-performance ports, with HP0/HP1 used for reads and HP2/HP3 used for writes. In contrast, the Zybo design targets the Zynq-7000 device \texttt{xc7z010clg400-1} and instantiates two accelerator engines using a 75 MHz PL clock. The Zybo version uses one AXI-GP control path and two HP data paths, where HP0 is used for reads and HP1 for writes.

\begin{table}[h]
\centering
\small
\caption{Post-synthesis resource utilization for the ZCU104 and Zybo \ac{SoC}-FPGA implementations.}
\label{tab:resource_utilization}
\begin{tabular}{@{}lcc@{}}
\toprule
\textbf{Resource} & \textbf{ZCU104, 12 engines} & \textbf{Zybo, 2 engines} \\
\midrule
LUTs & 59,228 / 230,400 (25.71\%) & 12,453 / 17,600 (70.76\%) \\
Registers & 50,349 / 460,800 (10.93\%) & 11,493 / 35,200 (32.65\%) \\
BRAM tiles & 270 / 312 (86.54\%) & 45 / 60 (75.00\%) \\
DSPs & 12 / 1,728 (0.69\%) & 2 / 80 (2.50\%) \\
Clock buffers & 1 / 96 (1.04\%) & 1 / 32 (3.13\%) \\
\bottomrule
\end{tabular}
\end{table}

The results show that the accelerator is mainly constrained by on-chip memory rather than arithmetic resources. This is expected because \ac{TM} training relies primarily on logical operations and automaton-state storage rather than dense floating-point computation. On the ZCU104, the larger programmable logic allows 12 engines to be instantiated while using only 25.71\% of LUTs and 10.93\% of registers, although BRAM utilization reaches 86.54\%. On the Zybo, the design remains feasible with two engines, but the smaller device is more tightly constrained, using 70.76\% of LUTs and 75.00\% of BRAM tiles. The low DSP utilization on both devices confirms that the architecture does not depend on conventional multiply-accumulate resources, supporting the logic-oriented design motivation of \ac{FastOmniTMAE}.
\end{document}